\documentclass[runningheads]{llncs}
\usepackage[T1]{fontenc}
\usepackage{graphicx}
\usepackage{comment}
\usepackage{multirow}
\usepackage{subcaption} 
\usepackage{booktabs}
\usepackage{microtype}
\usepackage{float}
\usepackage{amsmath}
\usepackage{amssymb}
\usepackage{latexsym}
\usepackage{pifont}
\usepackage[table,xcdraw]{xcolor}
\usepackage{tabularx}
\usepackage{caption}
\usepackage{stfloats}
\usepackage{url}
\usepackage{svg}

\usepackage{hyperref}
\usepackage{colortbl}
\definecolor{lightgray}{gray}{0.93}

\usepackage{color}

\begin{document}
\title{PaCoNet: Deep Data Extraction for Parallel Coordinates}
\author{Poonam Poonam\inst{1}\and
Hannah Kniesel\inst{1} \and
Pere-Pau Vázquez\inst{2} \and
Timo Ropinski\inst{1}}
\authorrunning{P. Poonam et al.}
\institute{Ulm University, Germany \inst{1} \\
\email{\{poonam.poonam, hannah.kniesel, timo.ropinski\}@uni-ulm.de}\\
Universitat Politècnica de Catalunya, Barcelona, Spain \inst{2}\\
\email{pere.pau.vazquez@upc.edu}
}
\maketitle              % typeset the header of the contribution
\begin{abstract}
Extracting data from visualizations has long challenged computer vision, with current research focused on bar, line, and pie charts, among other low-dimensional visualizations. However, parallel coordinates as a widely used high-dimensional data visualization approach, remain largely unexplored in this context. As parallel coordinate plots can quickly become cluttered and difficult to interpret when poorly designed or densely populated, automated data extraction from such visualizations is of particular interest. In this paper, we propose PaCoNet, the first approach for parallel coordinate data extraction. PaCoNet not only extracts line coordinates, but also enables the extraction of individual data samples for further analysis. Towards this end, we make the following contributions. We present the first deep learning approach tailored for parallel coordinate analysis, and demonstrate that it outperforms unadapted baselines by a significant margin.  
We further introduce a large-scale parallel coordinate dataset for training and testing. Together, these key contributions enable for the first time the automated analysis and redesign of parallel coordinate plots. PaCoNet thus lays the groundwork for complex visualization analysis, and further advances the intersection of computer vision and data visualization. All code, trained models, and data generation scripts are available at: https://github.com/poonam2308/PaCoNet

\keywords{Data Extraction \and Charts \and Deep Learning}
\end{abstract}
%

%%%%%%%%%%%%%%%%%%%%%%%%%%%%%%
%                            %
%        Introduction        %
%                            %
%%%%%%%%%%%%%%%%%%%%%%%%%%%%%%
\section{Introduction}\label{sec:introduction}
\begin{figure*}[t]
    \centering
    \includegraphics[width=0.32\textwidth,  trim=50 0 40 0, clip]{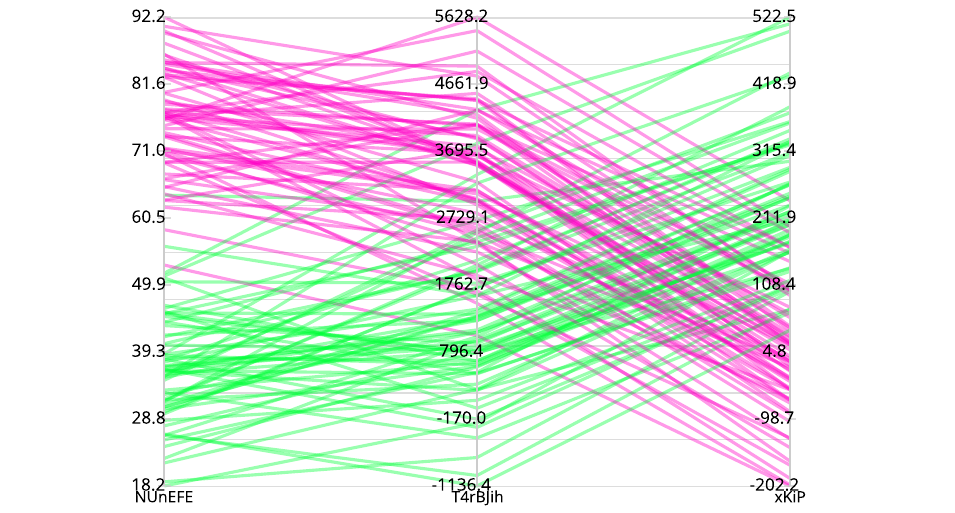}\hspace{-0.5em}
    \includegraphics[width=0.32\textwidth,  trim=50 0 35 0, clip]{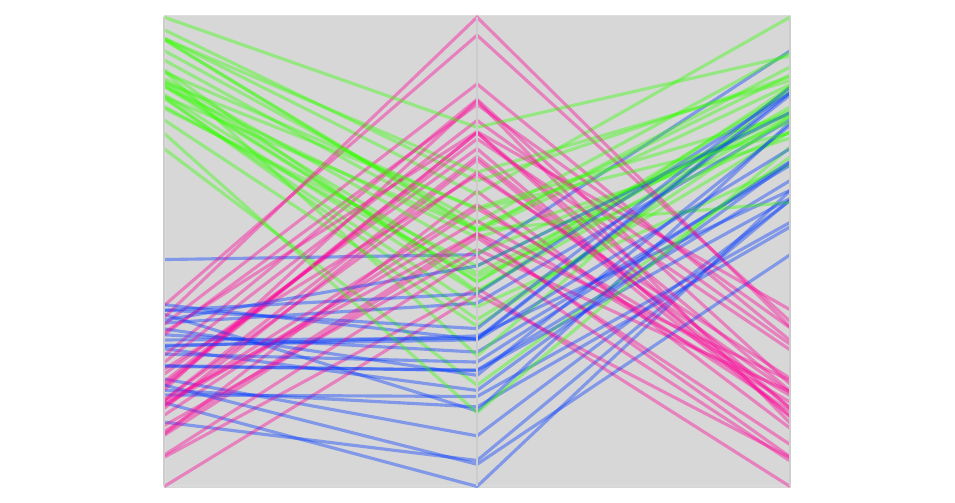}\hspace{-0.5em}
    \includegraphics[width=0.32\textwidth,  trim=35 0 40 0, clip]{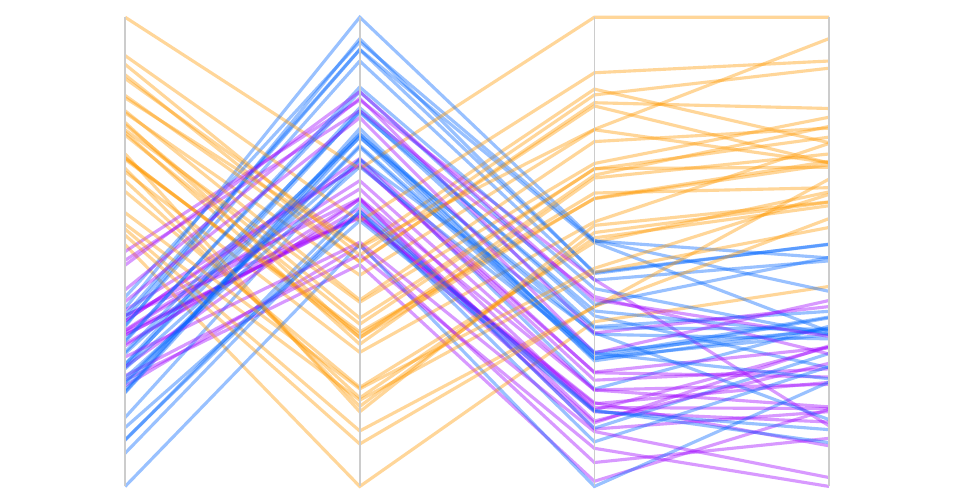}  \hspace{-0.5em}  
    
    \caption{%
      Example parallel coordinate plots extracted from our introduced training dataset. Parallel coordinates simultaneously depict multiple data dimensions within a single visualization, by exploiting parallel coordinate axis, such that different parameter configurations and feature distributions can be intuitively compared and interpreted.
    }
    \label{fig:teaser}
\end{figure*}

Data visualization plays a critical role in understanding complex, high-dimensional datasets.
Parallel coordinates, originally proposed by Inselberg in 1985~\cite{Inselberg1985}, visualize high-dimensional data by plotting each dimension along a separate parallel axis, whereby individual data samples are represented as polylines intersecting each axis at their corresponding values (see~\autoref{fig:teaser}).
Due to their benefits, parallel coordinates are widely employed for high-dimensional data analysis in domains such as finance~\cite{alsakran2010tile}, bioinformatics~\cite{boogaerts2012visualizing}, and engineering~\cite{kipouros2013parallel}. 
\\
Unfortunately, extracting data from a parallel coordinates plot is far more challenging than from simpler visualizations, such as bar charts or line
charts~\cite{Zhou2020,Lal2023,Shivasankaran2023}. Major challenges are the high density of overlapping lines and the inherent clutter this creates. Furthermore, in bar or line charts, each data point or category can often be individually distinguished, but in a parallel coordinates plot, numerous data dimensions and records are condensed into a single view, causing significant overdraw that obscures precise values. This is also demonstrated by the difficulties, modern vision-language models (VLMs)~\cite{chatgpt2025,team2023gemini} have to extract numerical values from parallel coordinates. As a result, while one may broadly infer data ranges or clustering patterns, extracting fine-grained, per-sample data currently requires direct access to the original dataset.
\\
Building on these challenges, we present PaCoNet, the first deep learning approach specifically tailored to extracting data from parallel coordinates. 
Our method not only retrieves crucial information but also enables to apply more complex analysis to the extracted data. To achieve these goals, PaCoNet has been designed to support quantitative analysis by extracting the geometric structure and relative data trajectories encoded in the plots. Besides proposing PaCoNet, we further release the first large-scale parallel coordinate training dataset, which is an essential requirement for further research in this direction. 
We quantitatively and qualitatively analyze PaCoNet, and demonstrate that it not only is the first deep parallel coordinate analysis approach, but that it also outperforms all previous baselines. Thus, within this paper, we make the following contributions:

\begin{itemize}
\item We present PaCoNet as the first deep learning approach for extracting data from parallel coordinates.
\item We introduce a large-scale parallel coordinate training and benchmark dataset.
\item We demonstrate PaCoNet's capabilities on real-world datasets, by extracting data and showcasing chart redesign, including axis reordering and recoloring.
\end{itemize}

%%%%%%%%%%%%%%%%%%%%%%%%%%%%%%
%                            %
%        Related Work        %
%                            %
%%%%%%%%%%%%%%%%%%%%%%%%%%%%%%
\section{Related Work}\label{sec:relatedwork}
\paragraph{Parallel Coordinates.} Parallel coordinates, introduced by Inselberg~\cite{Inselberg1985}, are a foundational visualization technique for high-dimensional data, mapping multidimensional samples to polylines intersecting parallel axes. Extensive prior work has focused on improving their visual interpretability through rendering strategies and interactive techniques~\cite{DBLP:conf/eurographics/HeinrichW13}. However, these efforts primarily target human interaction and visualization quality~\cite{Zhou2008,Tyagi2023} and highlight the active evolution of parallel coordinates visualization~\cite{lind2009many,wilks2018new}, rather than automated data extraction from rendered plots.
\paragraph{Deep Chart Analysis.}
Automated chart analysis has attracted significant attention, with early systems such as ReVision~\cite{Savva2011} focusing on chart classification and structural understanding. More recent deep learning approaches target precise numerical extraction, including methods for bar charts~\cite{Zhou2020} and line charts~\cite{Lal2023,Shivasankaran2023}, as well as comprehensive pipelines such as ChartOCR~\cite{Luo2021} that integrate detection and OCR.
Despite significant progress across bar, line, and pie charts~\cite{cui2024generalization,kato2022parsing,mustafa2023charteye,Liu2019}, parallel coordinates plots have not been previously addressed by deep learning approaches.  
Broader chart-mining pipelines, such as those developed in the ChartInfo and ChartQA benchmarks~\cite{davila2024chart,masry2022chartqa}, typically focus on recovering semantic information by parsing textual elements, axis scales, and legends. In contrast, PaCoNet targets the geometric extraction of dense polyline structures in parallel coordinates plots, which pose unique challenges due to heavy overdraw and frequent crossings and are not explicitly addressed in these benchmarks.

\paragraph{Deep Line Detection.} 
Line detection is a fundamental computer vision task that has advanced significantly with deep learning. Beyond classical approaches such as the Hough Transform~\cite{hough1962method}, modern methods incorporate global reasoning through learnable Hough-like formulations~\cite{Han2020DeepHT,Lin2020}, transformer-based architectures~\cite{Xu2021}, and holistic wireframe parsing~\cite{Xue2020}.
General-purpose line and segment detectors and classical Hough-based approaches~\cite{hough1962method,Han2020DeepHT} are primarily designed to detect sparse, isolated line structures in natural images. In contrast, parallel coordinates plots contain extremely dense, overlapping, and highly intersecting polylines, where these methods tend to produce fragmented or noisy detections that are difficult to associate into coherent data trajectories.

Among existing approaches, DHLP~\cite{Lin2020} is well suited for extracting lines from parallel coordinate plots due to their structural regularity. Such plots consist of densely overlapping, predominantly straight line segments that often span the full vertical extent of the image. DHLP incorporates global geometric priors and aggregates evidence across entire line extents in a Hough-transform–inspired manner, making it robust to dense crossings, anti-aliasing artifacts, and low-contrast regions. As a result, DHLP provides structured line representations that can be naturally adapted to recover line coordinates in this domain.

%%%%%%%%%%%%%%%%%%%%%%%%%%%%%%
%                            %
%         Data Sets          %
%                            %
%%%%%%%%%%%%%%%%%%%%%%%%%%%%%%
\section{Dataset Construction}
\label{sec:dataset}
Training models for automatic data extraction from parallel coordinates plots requires large amounts of annotated data, which are not publicly available. To address this limitation, we construct a large-scale synthetic dataset whose design is informed by statistical properties observed in real-world parallel coordinates plots. The later we have carefully collected and curated to form a real-world parallel coordinate data set.
We first summarize this real-world dataset used for statistical analysis, and then describe the synthetic data generation pipeline guided by these statistics.

\begin{figure*}[t]
    \centering
    % -------- First row --------
    \includegraphics[width=0.32\textwidth]{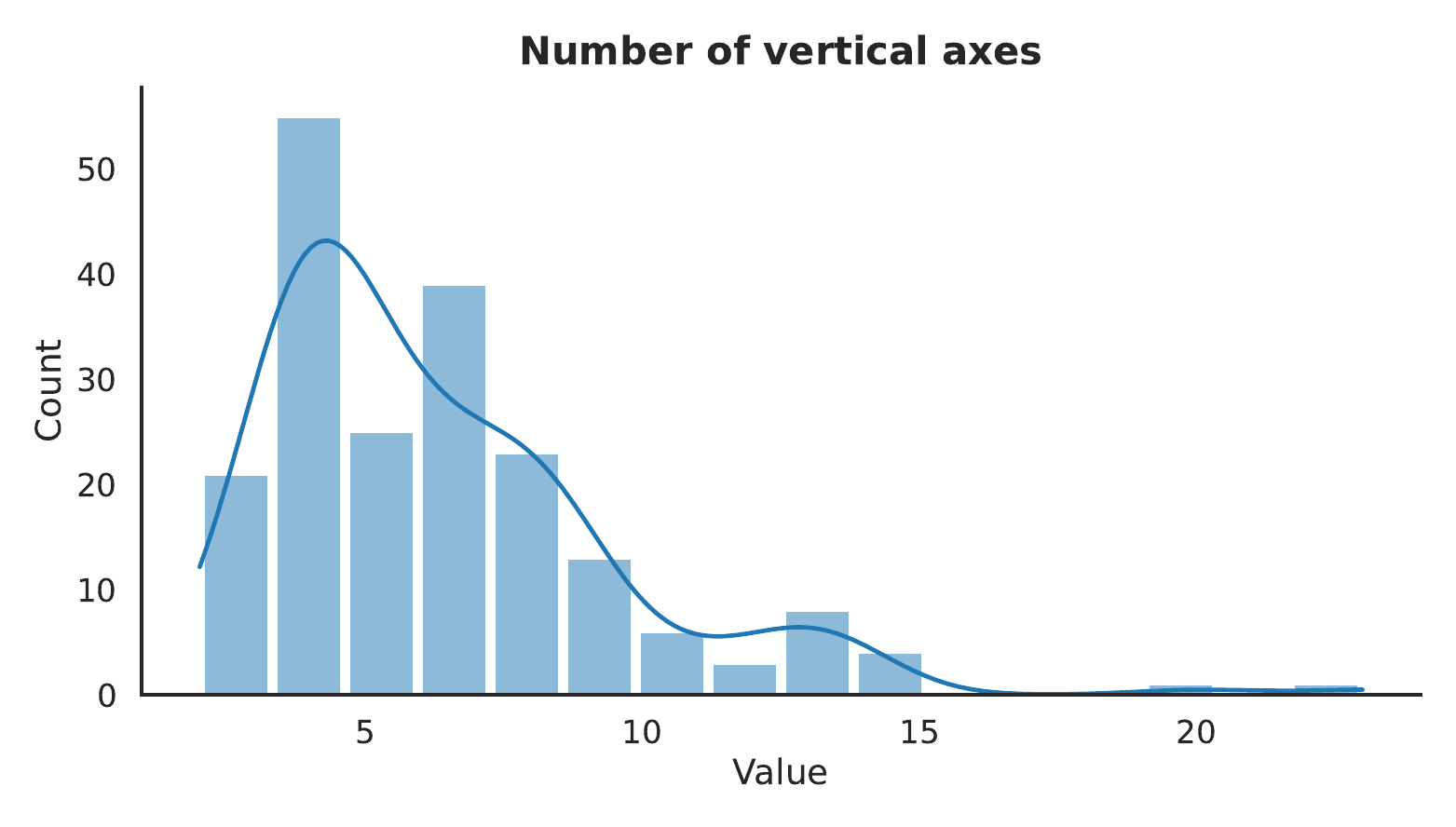}
    \includegraphics[width=0.32\textwidth]{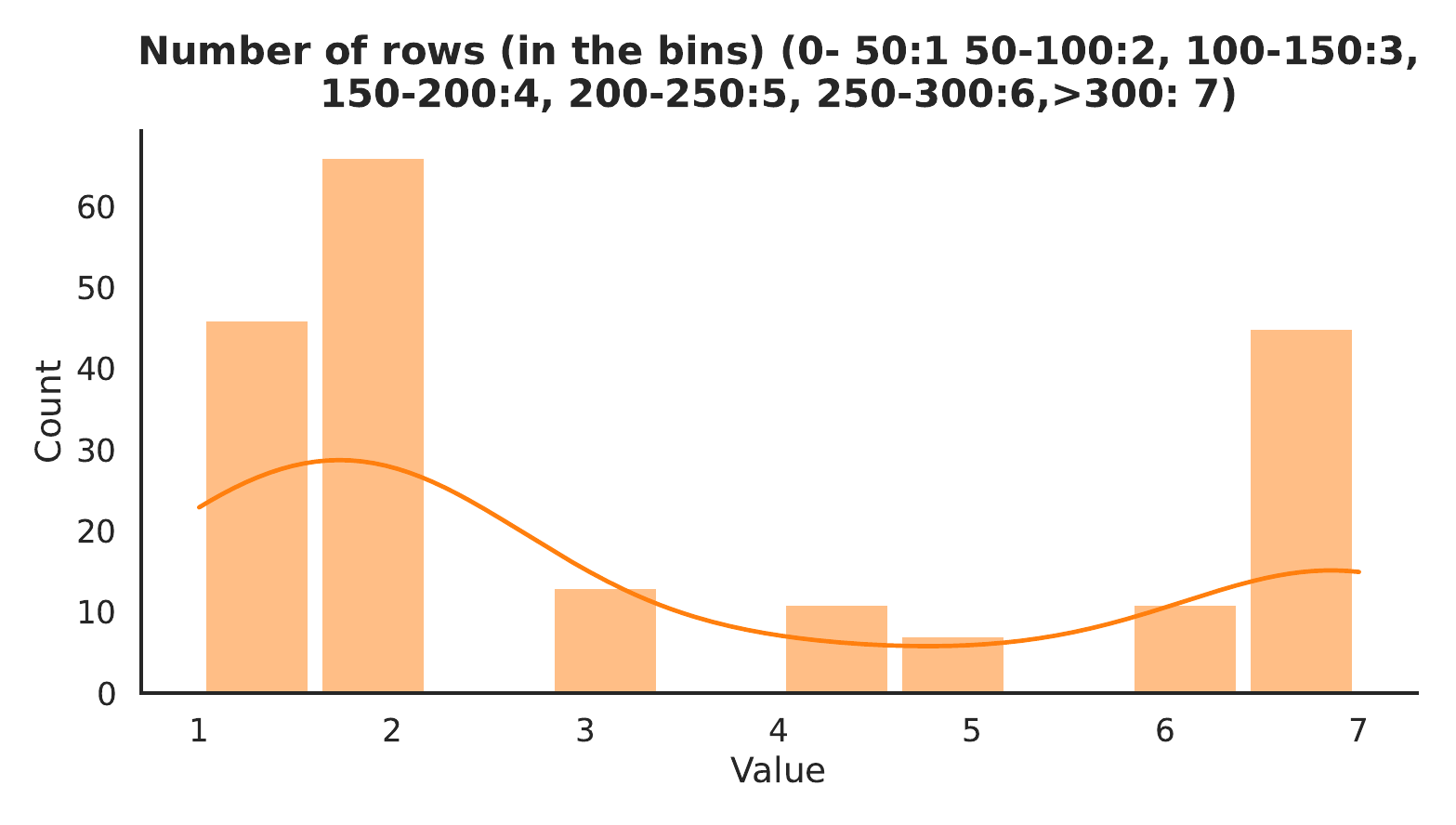}
    \includegraphics[width=0.32\textwidth]{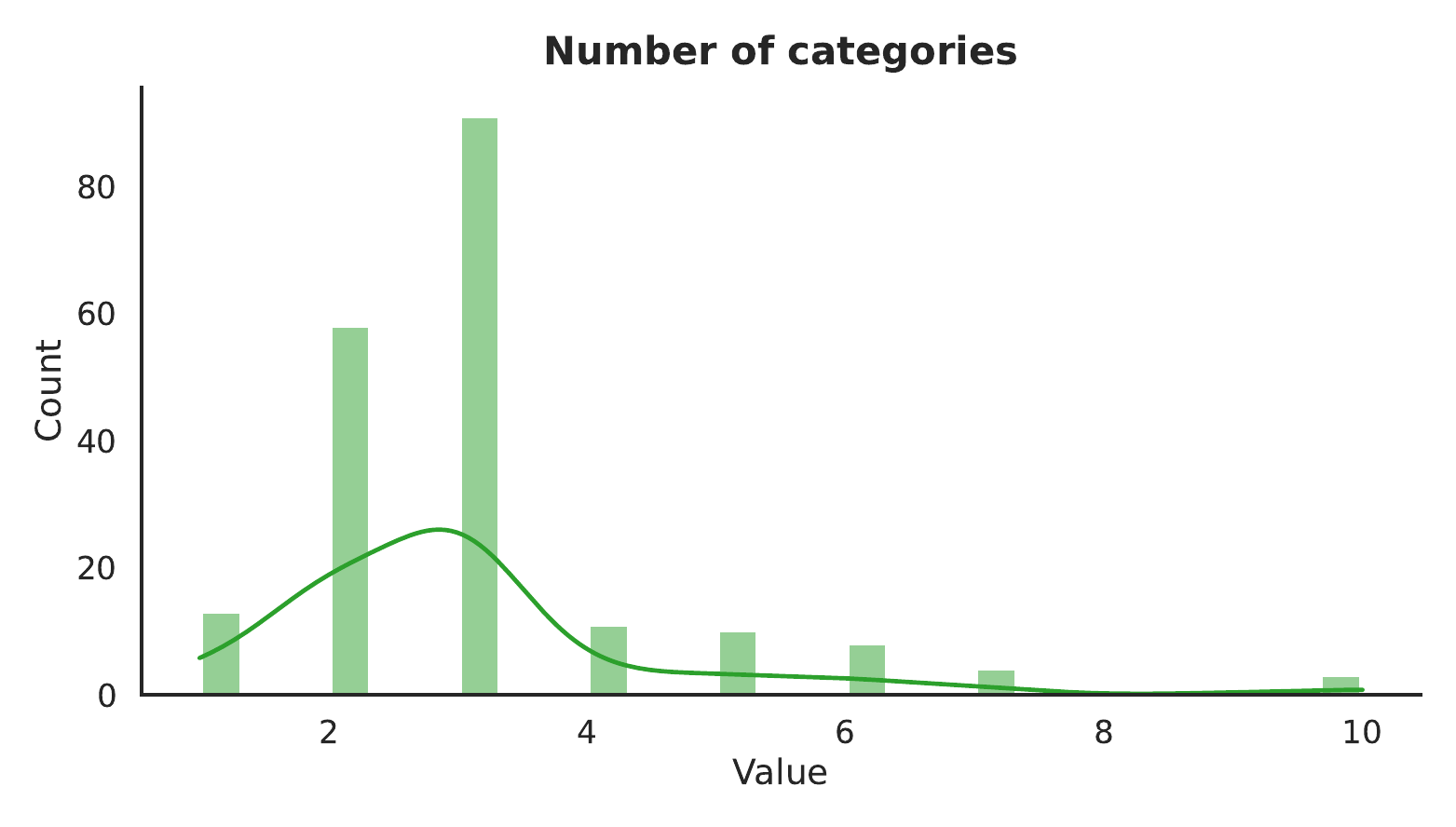}

    \vspace{0.3cm}

    % -------- Second row --------
    \includegraphics[width=0.32\textwidth]{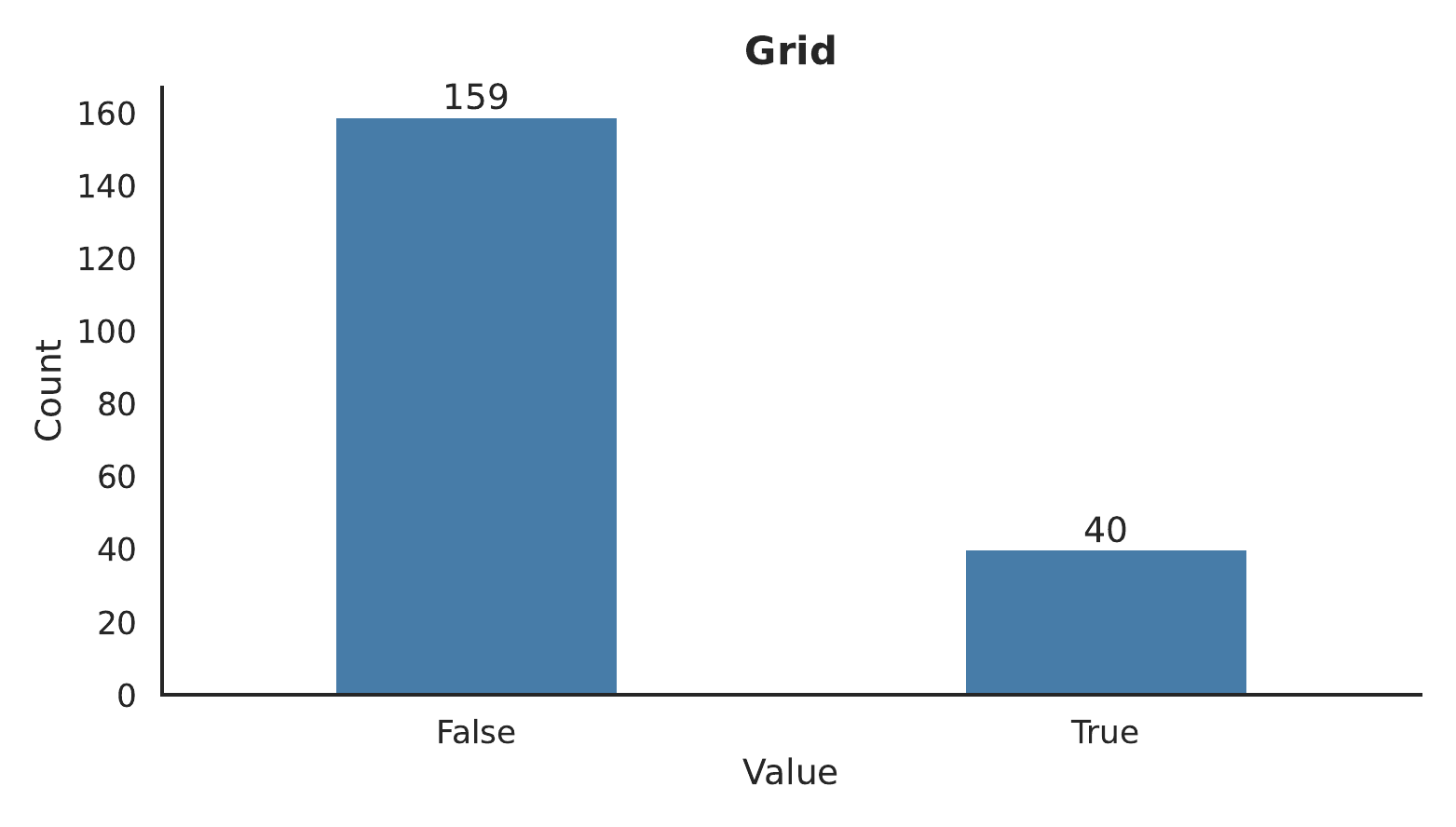}
    \includegraphics[width=0.32\textwidth]{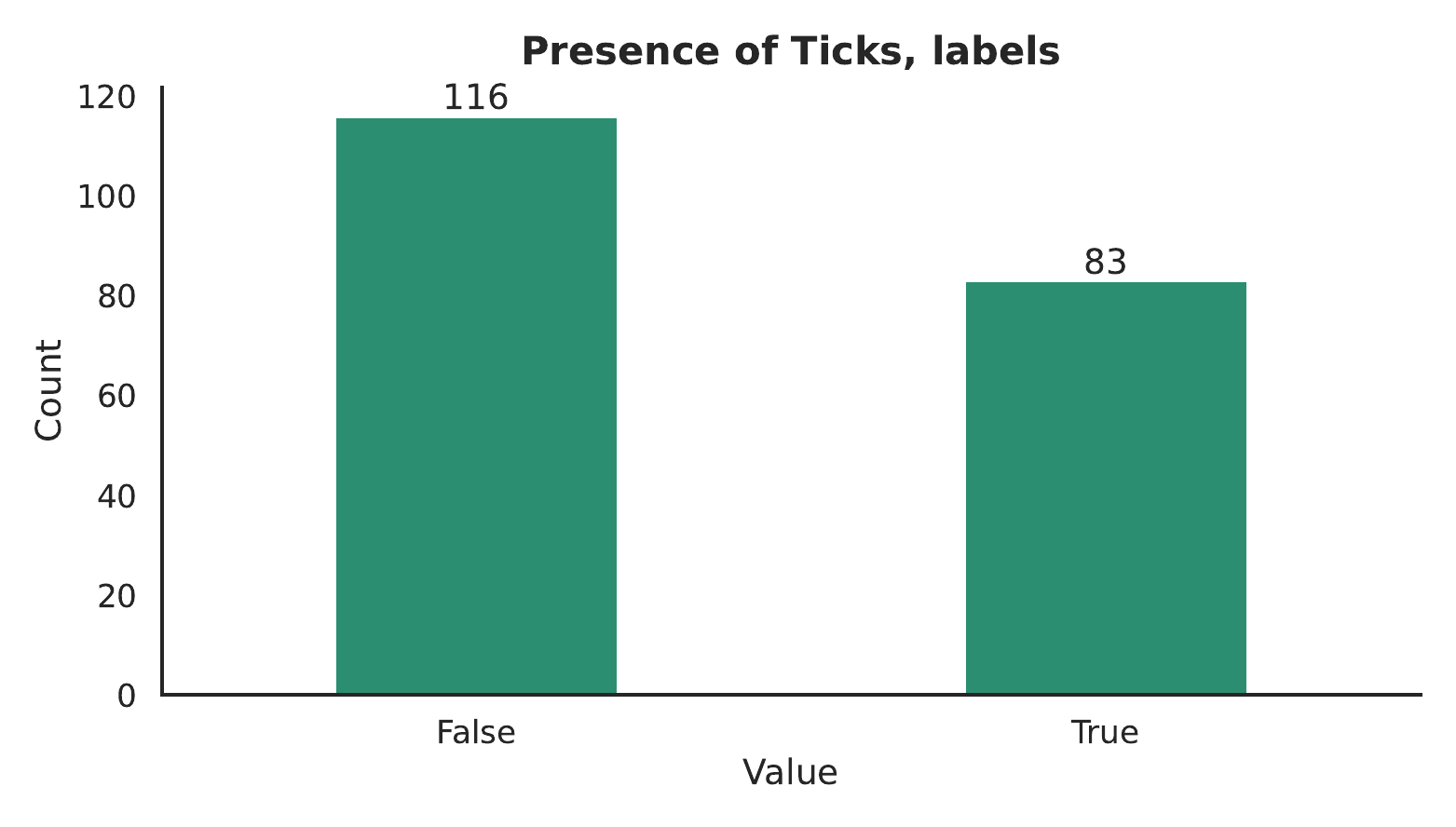}
    \includegraphics[width=0.32\textwidth]{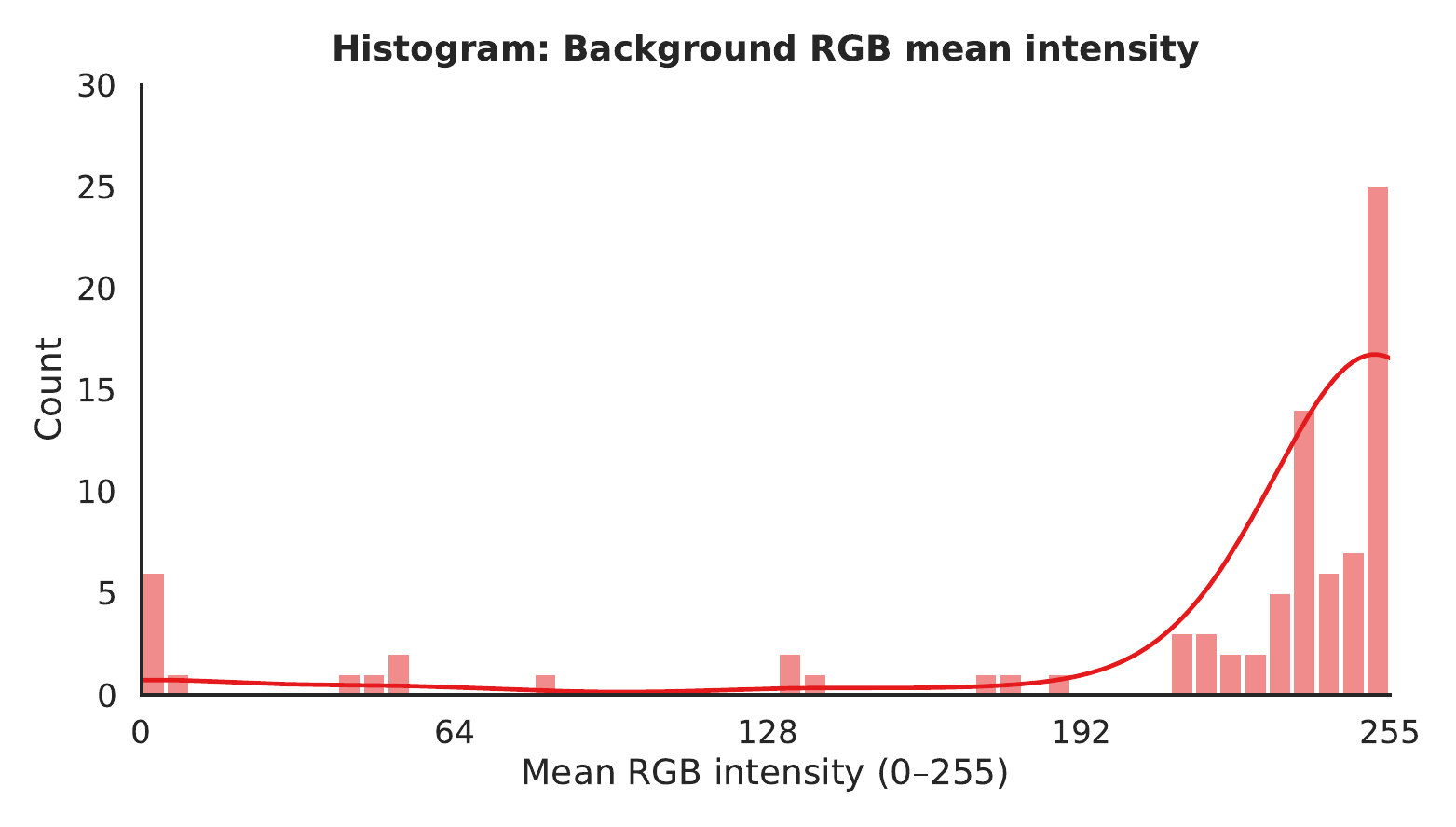}

    \caption{Distribution of key visual and structural properties in the real-world chart dataset. Top row (left to right) shows the distributions of the number of vertical axes, the number of rows (binned), and the number of categories. Bottom row shows the distribution of grid presence, tick presence, label presence, and the background RGB mean intensity.}
    \label{fig:realworldstats}
\end{figure*}

\subsection{Real-World Data}
\label{subsec:realworlddata}
To obtain representative real-world parallel coordinates plots, we collect and curate a set of approximately 200 images from publicly available sources, ensuring that the dataset reflects realistic and interpretable examples encountered in practice. 
We analyze the real-world dataset to estimate key visual and structural properties of parallel coordinates plots, including the number of axes, data density, categorical color usage, grid and label presence, and background appearance. The resulting distributions, summarized in~\autoref{fig:realworldstats}, are used to guide the parameter ranges and design decisions in our synthetic data generation process. Detailed dataset collection procedures and statistics extraction protocols are provided in the supplementary material.

\subsection{Synthetic Data Generation}
\label{subsec:syntheticdata}
Due to its limited size and lack of ground-truth annotations, the real-world dataset cannot be used directly for supervised learning. We therefore generate a large synthetic dataset of parallel coordinates plots with exact ground-truth annotations.
The design of the synthetic data generator is informed by the statistical analysis of the real-world dataset (~\autoref{fig:realworldstats}). Rather than generating plots with arbitrary parameters, we sample key properties from distributions observed in real-world visualizations, ensuring realistic variation while retaining full control and precise annotations.
For each synthetic plot, we sample the number of axes, data rows, and categories from empirical distributions estimated from the real-world dataset, ensuring coverage of both sparse and densely populated plots. In addition, we model key visual attributes affecting extraction performance, including grid lines, ticks, labels, background brightness, and line colors, by sampling from observed real-world distributions.

\paragraph{Data Generation and Annotation.}
Using the sampled parameters, we generate parallel coordinates plots programmatically using the Vega-Lite visualization grammar~\cite{satyanarayan2016vega} and render them as raster images a fixed
resolution of 600 × 300 pixels. Since the underlying data values and plot layout are known by construction, we obtain exact ground-truth annotations for each plot, including axis locations, data values, and polyline correspondences.
This process yields a large, diverse, and fully annotated synthetic dataset. We use 5{,}000 images for training and validation ($80\%/20\%$ split) and an additional 1{,}000 images for testing. Additional implementation details of the synthetic data generator are provided in the supplementary material.

%%%%%%%%%%%%%%%%%%%%%%%%%%%%%%
%                            %
%           Method           %
%                            %
%%%%%%%%%%%%%%%%%%%%%%%%%%%%%%
\section{Method}\label{sec:method}
\begin{figure*}[t]
  \centering
  \includegraphics[width=1.0\linewidth]{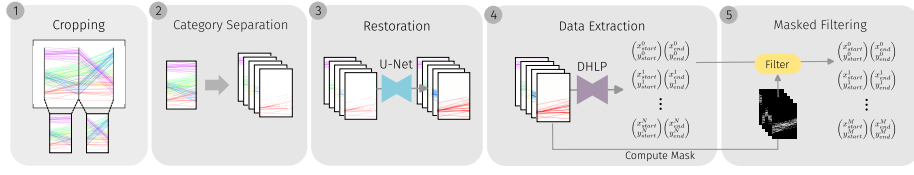}
  \caption{Overview of the proposed PaCoNet framework for data extraction from parallel coordinates plots. The pipeline consists of five stages: 
(1) \textbf{Cropping}, where the input plot is partitioned into regions between adjacent vertical axes; 
(2) \textbf{Category Separation}, which separates individual line categories to mitigate overdraw; 
(3) \textbf{Restoration}, using a U-Net-based network to enhance line continuity and suppress artifacts; 
(4) \textbf{Data Extraction}, where a deep neural network (DHLP)~\cite{Lin2020} predicts polyline coordinates; and 
(5) \textbf{Masked Filtering}, which applies mask-based filtering to refine the final set of extracted coordinates.}
   \label{fig:workflow}
\end{figure*}

Our pipeline for data extraction from parallel coordinates plots is designed to address challenges such as visual clutter, line overdraw, and multi-category representations. The method comprises a sequence of modular steps that progressively reduce ambiguity and improve line separability. First, the parallel coordinates plot is cropped at the vertical axes, next we separate individual categories by their color. Since this separation may introduce artifacts, we subsequently apply a line restoration step. The coordinates are then extracted using a combination of learning-based and algorithmic components. Finally, a postprocessing stage suppresses falsely identified lines. An overview of the workflow is shown in~\autoref{fig:workflow}. Throughout this section, we refer to a category as a group of polylines sharing the same visual color encoding.

\subsection{Cropping}
To reduce task complexity, we detect vertical axes and crop the plot into regions between adjacent axes, each corresponding to the space where polylines transition between dimensions (~\autoref{fig:workflow}(1)).
For synthetic data, axis positions are obtained directly from the generation parameters, while for real-world images vertical axes are detected using a lightweight geometric procedure based on vertical line detection and projection-profile analysis.
This allows subsequent processing stages to operate on localized line segments with no interference from unrelated axes. Although extraction is performed on crops between adjacent axes, polyline identities are preserved globally by associating corresponding segments across successive axis pairs based on axis order and spatial continuity.

\subsection{Category Separation}
\label{cat_sep_tech}
Parallel coordinates plots often encode multiple categories using color, which exacerbates visual clutter and overdraw as the number of categories increases. To reduce this complexity, we introduce an explicit category separation step (~\autoref{fig:workflow}(2)).
As preprocessing, we remove non-essential visual elements such as axis labels, ticks, and grid annotations, retaining only the colored polylines.
Since category information is not explicitly encoded, we infer category structure from color distributions using two strategies: \textbf{peak-based separation} and \textbf{clustering-based separation}. Both operate without prior knowledge of the number of categories and are robust to low color contrast.
This formulation assumes categories are encoded using discrete, stable color hues, reflecting common practice in categorical parallel coordinates plots; grayscale, monochrome, or continuously varying colormaps are outside the scope of this work.

\paragraph{Peak-Based Separation.} 
To estimate the number of categories $N$, we compute a one-dimensional histogram over all hue values in the image. Let $h_i \in [0,1]$ denote the hue value of pixel \(i\), and let
$$
H(k) = \sum_i 1\bigl(h_i \in [b_k, b_{k+1})\bigr)
$$
be the histogram count of bin $k$ with bin boundaries $\{b_k\}$. Then, we apply peak detection to identify a set of local maxima $\{p_1, \dots, p_N\}$, where each peak corresponds to a dominant color mode in the plot and thus defines one category.
Given the detected peaks, each pixel is assigned to the nearest peak in hue space,
$$
c_i = \arg\min_{j \in \{1,\dots,N\}} \lvert h_i - p_j \rvert,
$$
thereby grouping pixels with similar colors into a common category. Using this assignment, the original parallel coordinates plot is decomposed into $N$ category-specific images by retaining only pixels belonging to a given category and suppressing all others.
As a result, we obtain $N$ cropped images, one for each inferred category.

\paragraph{Clustering-Based Separation.}
In addition to peak detection, we investigate clustering as an alternative mechanism for category separation that directly assigns pixels or line fragments to category groups. As before, clustering is performed on the hue component in HSV color space. Since the number of categories present in a parallel coordinates plot is unknown a priori, density-based clustering methods are particularly well suited to this task. 
Specifically, we employ DBSCAN~\cite{ester1996density} and HDBSCAN~\cite{mcinnes2017hdbscan}, which identify clusters as dense regions in feature space without requiring the number of clusters to be specified in advance. Given hue values $\{h_i\}$, DBSCAN groups samples based on a neighborhood radius $\varepsilon$ and a minimum number of samples $m$. A pixel $i$ is considered a core point if
$$
\lvert \{\, j \mid \lvert h_i - h_j \rvert \le \varepsilon \,\} \rvert \ge m,
$$
and clusters are formed by connecting density-reachable core points. Points that are not assigned to any cluster are treated as noise.
The resulting clusters correspond to the inferred categories and are used to decompose the original plot into category-specific images. These strategies define two variants of our method, PaCoNet$_{Peak}$ and PaCoNet$_{DBScan}$.

\subsection{Restoration}\label{image_restoration} 
Due to visual clutter and strong overdraw, particularly in parallel coordinate plots with many categories, the previous category separation step can introduce artifacts, resulting in noisy lines (see~\autoref{fig:results}, first column) and complicating subsequent analysis.
These imperfections primarily arise from cluster and peak-based separation methods, where overlapping lines, anti-aliasing effects, and compression artifacts can produce broken polylines, spurious gaps, color bleeding between categories, and isolated noise pixels.
\\
To address these issues, we integrate a restoration step using a UNet-based neural network~\cite{Ronneberger2015UNetCN}, as illustrated in~\autoref{fig:workflow}(3). Let \(I_s\) denote the separated, artifact-prone image for a given category, and let \(I_c\) denote the corresponding clean image derived from our synthetic ground truth. The UNet is trained to learn a mapping
$$
f_\theta: I_s \mapsto I_c,
$$
where $\theta$ represents the network parameters. 
The training objective is to minimize a pixel-wise mean squared error (MSE) loss:
$$
\mathcal{L}(\theta) = \frac{1}{|P|} \sum_{p \in P} \lVert f_\theta(I_s)_p - I_{c,p} \rVert_2^2,
$$
where $P$ denotes the set of pixels and $\lVert \cdot \rVert_2$ denotes the $\ell_2$ norm.
After training, the U-Net restores noisy lines and corrects visual artifacts, producing clean, category-specific plots suitable for downstream analysis.

\subsection{Data Extraction} 
With the now category-separated and restored image segments, we proceed to detect the individual data lines and extract data using DHLP~\cite{Lin2020}. DHLP is particularly well suited for parallel coordinates plots, as discussed in~\autoref{sec:relatedwork}, because it embeds global geometric priors that allow robust detection of densely overlapping, mostly straight lines, even in the presence of low-contrast regions. Global polylines are reconstructed by linking extracted line segments across adjacent axis crops in sequence, yielding a single continuous polyline per category that spans all axes. The network is built upon a fully convolutional architecture, and to ensure consistency and reproducibility, we follow the same network architecture and training setup as described in~\cite{Lin2020}. This step (~\autoref{fig:workflow}(4)) is pivotal for accurately identifying and extracting the lines, which is crucial for reconstructing the underlying data samples from parallel coordinate charts. The resulting data representation captures relative positions and trends across axes in image coordinates, without requiring explicit parsing of axis ticks, labels, or numeric scales.

\subsection{Masked Filtering}\label{masking}
Finally, we refine the predicted lines using our proposed masked filtering (~\autoref{fig:workflow}(5)). After restoration with the UNet, we obtain a set of cleaned, category-specific images, denoted as \(\{I_c^{(k)}\}_{k=1}^{N}\), where \(N\) is the number of categories. Each restored image is then binarized to generate a mask highlighting the detected lines:
$$
M^{(k)}(x,y) = 
\begin{cases}
1, & \text{if } I_c^{(k)}(x,y) \ge \tau,\\
0, & \text{otherwise,}
\end{cases}
$$
where $I_c^{(k)}(x,y)$ is the pixel intensity at position $(x,y)$ in category $k$ and $\tau$ is a binarization threshold.
\\
Let $\mathcal{L} = \{L_1, L_2, \dots, L_m\}$ denote the set of lines detected by DHLP in the restored images. Each line $L_i$ is represented as a sequence of pixel coordinates $\{(x_{i,j},y_{i,j})\}_{j=1}^{n_i}$. To enforce spatial consistency with the restored mask, we retain only those lines for which a sufficient fraction of pixels lies within the mask:
$$
\hat{\mathcal{L}} = \left\{ L_i \in \mathcal{L} \;\Big|\; \frac{1}{n_i} \sum_{j=1}^{n_i} M^{(k)}(x_{i,j},y_{i,j}) \ge \alpha \right\},
$$
where $\alpha \in [0,1]$ is a threshold specifying the minimum fraction of pixels that must lie inside the mask for the line to be retained. Lines not satisfying this criterion are removed.
\\
This masked filtering step ensures that only lines corresponding to valid, restored regions are kept, effectively suppressing spurious detections outside the category-specific regions. The output $\hat{\mathcal{L}}$ forms the final set of lines used for reconstruction and analysis.

%%%%%%%%%%%%%%%%%%%%%%%%%%%%%%
%                            %
%        Experiments         %
%                            %
%%%%%%%%%%%%%%%%%%%%%%%%%%%%%%
\section{Experiments}\label{sec:experiments}
We evaluate PaCoNet on both synthetic and real-world datasets. Since no prior method addresses end-to-end data extraction from parallel coordinates plots, our experiments validate feasibility, analyze key design choices, and assess robustness under varying visual conditions. Quantitative evaluation is restricted to synthetic data, where precise ground truth is available, while real-world results are assessed qualitatively.
\begin{figure*}[t]
  \centering
  \includegraphics[width=0.7\linewidth]{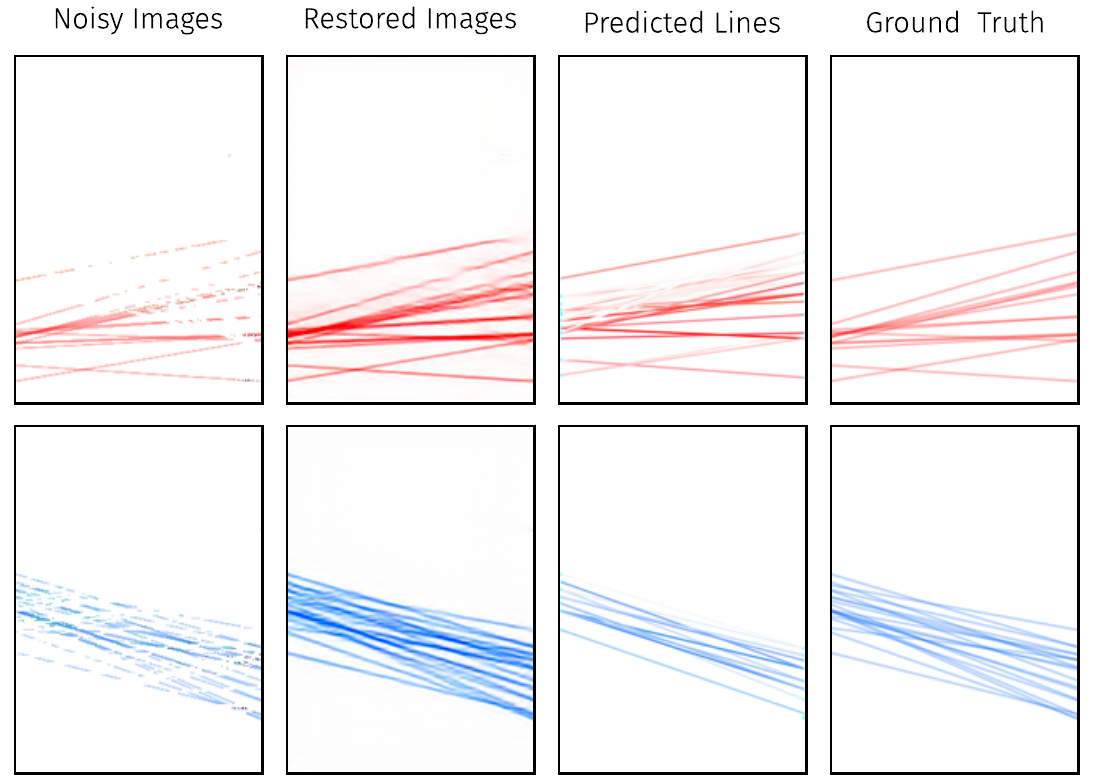}
  \caption{Qualitative results on the synthetic test set before chart reconstruction. From left to right: noisy input images, restored images, predicted line structures, and ground truth. The results highlight the effectiveness of restoration and line prediction.}
   \label{fig:results}
\end{figure*}

\subsection{Experimental Setup}\label{exp_setup}
\paragraph{Evaluation Protocol.}
All models are trained exclusively on synthetically generated parallel coordinate plots with complete ground-truth annotations. Training is performed on NVIDIA A6000 and A100 GPUs, with each model requiring approximately 10–12 hours. We adopt a standard U-Net~\cite{Ronneberger2015UNetCN} with established training hyperparameters for the restoration network and retain the hyperparameter configuration as in the original DHLP paper~\cite{Lin2020}. 
We evaluate PaCoNet quantitatively and qualitatively on the synthetic test set, which reflects real-world plot distributions (see~\autoref{sec:dataset}), and additionally report qualitative results on real-world examples where ground-truth annotations are unavailable.

\paragraph{Evaluation Metrics.}\label{eval_metric} 
We use two complementary metrics. We use Mean Absolute Error~\cite{willmott2005advantages} to quantify the difference between the number of predicted and ground-truth lines. As we similarly use MAE to quantify multiple types of error, we refer to this metric as "LC-MAE". We additionally report structural Average Precision (sAP)~\cite{Lin2020}, which measures detection accuracy under geometric tolerance thresholds. We report sAP at AP$^{5}$ and AP$^{10}$.
\paragraph{Category Separation Metrics.} \label{cat_eval_metric}
To evaluate category separation design choices in ablation studies, we additionally report the Silhouette Coefficient (SC)~\cite{lovmar2005silhouette} to assess color cluster separability, category-count MAE (CC-MAE)~\cite{willmott2005advantages} to measure errors in predicted category counts, and Intersection-over-Union (IoU)~\cite{Rezatofighi_2019_CVPR} to quantify spatial agreement between predicted and ground-truth category masks when pixel-level assignments are available. Additional details are provided in the supplementary material.

\subsection{Main Results}
Our results in~\autoref{tab:combined_results} demonstrate better performance compared to the baseline of VLMs, DHT, and DHLP respectively.
We compare PaCoNet against both general-purpose and task-specific baselines. As general-purpose baselines, we evaluate VLMs, including OpenAI models~\cite{chatgpt2025} and Gemini~\cite{team2023gemini}, which reflect current user practice for chart understanding.
Although these models do not natively output explicit line-level predictions, we prompt them to regress line coordinates on cropped regions enabling a quantitative comparison using LC-MAE. As a task-specific baseline, we compare against DHT~\cite{Han2020DeepHT} and DHLP~\cite{Lin2020}, which are applied independently to each cropped region and evaluated using sAP. 
\begin{table}[t]
\centering
\caption{Quantitative evaluation of PaCoNet on the synthetic test set.
\textbf{(a)} LC-MAE comparison against VLMs, DHT and DHLP baseline.
\textbf{(b)} sAP comparison against DHT and DHLP evaluating line detection. Best results are highlighted in bold.}
\label{tab:combined_results}
\begin{subtable}[t]{0.29\linewidth}
\centering
\caption{LC-MAE $\downarrow$}
\resizebox{\linewidth}{!}{%
\begin{tabular}{lc}
\toprule
Method & LC-MAE \\
\midrule
GPT4.1~\cite{chatgpt2025}  & 0.80\\
GPT5.2~\cite{chatgpt2025}  & 0.61\\
Gemini2.5-flash~\cite{team2023gemini} & 0.71 \\
Gemini3.0-flash~\cite{team2023gemini} & 0.68 \\
DHT~\cite{Han2020DeepHT} & 0.95\\
DHLP~\cite{Lin2020} & 0.47 \\
\midrule
PaCoNet$_{Peak}$ & 0.38 \\
PaCoNet$_{DBScan}$ & \textbf{0.37} \\
\bottomrule
\end{tabular}
}
\end{subtable}
\hfill
\begin{subtable}[t]{0.70\linewidth}
\centering
\caption{sAP $\uparrow$}
\resizebox{\linewidth}{!}{%
\begin{tabular}{lcccc}
\toprule
\multirow{2}{*}{Method} & 
\multicolumn{2}{c}{No Masked Filtering} & 
\multicolumn{2}{c}{Masked Filtering} \\
 & sAP$^5$ & sAP$^{10}$ & sAP$^5$ & sAP$^{10}$ \\
\midrule
DHT~\cite{Han2020DeepHT}  & \cellcolor{lightgray}0.96 & \cellcolor{lightgray}2.69 
& \cellcolor{lightgray}0.14 & \cellcolor{lightgray}0.34 \\
DHLP~\cite{Lin2020}  & \cellcolor{lightgray}28.00 & \cellcolor{lightgray}35.89 
& \cellcolor{lightgray}40.56 & \cellcolor{lightgray}46.55 \\
PaCoNet$_{Peak}$     & 40.48 & 44.24 & 61.66 & \textbf{68.39} \\
PaCoNet$_{DBScan}$   & 39.82 & 43.38 & 56.10 & 61.80 \\
\bottomrule
\end{tabular}
}
\end{subtable}
\end{table}

\subsection{Ablations} 
To assess the contribution of each component, we conduct ablation studies evaluating category separation, UNet-based restoration, and post-processing on line detection performance.

\paragraph{Category Separation.} 
We analyze pixel color distributions within each cropped region and compare RGB, LAB, and HSV color spaces for their ability to separate categorical color modes.
\autoref{tab:ablations_clustering_overview}(a) shows the benefits of separating the clusters using the hue component of the HSV color space.  
We further analyze the effect of input resolution and observe that clustering-based separation benefits from downscaled images, while peak-based separation performs best at full resolution (~\autoref{tab:ablations_clustering_overview}(b,c)).
We additionally compare DBSCAN~\cite{ester1996density} and HDBSCAN~\cite{mcinnes2017hdbscan}, finding that DBSCAN consistently yields better performance in our setting. This aligns with the one-dimensional hue feature space, where category colors form relatively uniform, well-separated clusters (~\autoref{tab:ablations_clustering_overview}(d)). 
\paragraph{Image Restoration.} 
In this ablation study, we examine the importance of image restoration. Specifically, we compare DHLP~\cite{Lin2020} trained on noisy images (see~\autoref{fig:results}, first column), due to the category separation, to DHLP~\cite{Lin2020} trained on the restored (see~\autoref{fig:results}, second column) images. 
Note that, we train separate UNet models for each category separation method: one for \textbf{peak-based separation} and one for \textbf{clustering-based separation}. In each case, the input $I_s$ is the image resulting from the respective separation method, and the target $I_c$ is its corresponding clean counterpart generated from synthetic ground truth images. 
The results, presented in~\autoref{tab:ablations}, demonstrate that training on denoised images enhances performance. The structured patterns in the images become more distinct after the denoising step, leading to more accurate line detections. For qualitative results please see~\autoref{fig:results}.
\paragraph{Masked Filtering.}  
We also assess the effectiveness of our masked filtering approach, as outlined in~\autoref{masking}. By applying this post-processing step, our evaluation yielded the best results, as demonstrated in~\autoref{tab:ablations}. These findings highlight the significance and efficiency of our masked filtering strategy.
\\
These ablation studies confirm that our proposed approach, which explicitly reduces clutter and overdraw prior to analysis by combining category-wise separation with UNet-based restoration, significantly improves line detection performance in parallel coordinate plots.
\begin{table}[t]
\centering
\caption{
Ablation studies on category separation:
(a) Comparison of color spaces for category separation using SC.
(b) Influence of input image resolution on cluster-based separation accuracy.
(c) Comparison of density-based clustering algorithms.
(d) Influence of input image resolution on peak-based category separation.
}
\label{tab:ablations_clustering_overview}
\resizebox{0.7\linewidth}{!}{%
\begin{tabular}{ccc}
% -------------------- Row 1 --------------------
\begin{minipage}{0.45\linewidth}
\centering
\textbf{(a)}\\[0.5ex]
\begin{tabular}{lc}
\toprule
Color Space & SC $\uparrow$ \\
\midrule
RGB       & 0.58 \\
LAB       & 0.61 \\
HSV$_H$   & \textbf{0.84} \\
HSV$_{Full}$ & 0.60 \\
\bottomrule
\end{tabular}
\end{minipage}
&
\begin{minipage}{0.45\linewidth}
\centering
\textbf{(b)}\\[0.5ex]
\begin{tabular}{lcc}
\toprule
Resolution & CC-MAE $\downarrow$ & IoU $\uparrow$ \\
\midrule
Full & \textbf{0.077} &0.33\\
Downscaled & 3.14 & \textbf{0.34}\\
\bottomrule
\end{tabular}
\end{minipage}
\\[2ex]
% -------------------- Row 2 --------------------
\begin{minipage}{0.45\linewidth}
\vspace{.3em}
\centering
\textbf{(c)}\\[0.5ex]
\begin{tabular}{lcc}
\toprule
Resolution & CC-MAE $\downarrow$ & IoU $\uparrow$\\
\midrule
Full & 4.27 & 0.35\\
Downscaled & \textbf{1.65}&  \textbf{0.36}\\
\bottomrule
\end{tabular}

\end{minipage}
&
\begin{minipage}{0.45\linewidth}
\centering
\textbf{(d)}\\[0.5ex]
\begin{tabular}{lcc}
\toprule
Method & CC-MAE $\downarrow$  &  IoU $\uparrow$\\
\midrule
DBSCAN~\cite{ester1996density} & \textbf{0.81} & \textbf{0.35}\\
HDBSCAN~\cite{mcinnes2017hdbscan} & 0.89  & \textbf{0.35}\\
\bottomrule
\end{tabular}
\end{minipage}
\end{tabular}
}
\end{table}

\begin{table}[t]
\centering
\caption{Ablation study assessing the impact of key components in our pipeline, including category separation, line restoration, and post-processing. sAP$^5$ and sAP$^{10}$ are reported.}
\label{tab:ablations}
\resizebox{.7\linewidth}{!}{%
\begin{tabular}{p{1.7cm} c c c c c}
\toprule
Category Separation & Restoration 
& \multicolumn{2}{c}{No Masked Filtering} 
& \multicolumn{2}{c}{Masked Filtering} \\
& 
& sAP$^5 \uparrow$ & sAP$^{10} \uparrow$ 
& sAP$^5 \uparrow$ & sAP$^{10} \uparrow$ \\
\midrule
Peaks & \ding{55} 
& \cellcolor{lightgray}33.49 & \cellcolor{lightgray}38.51 
& 36.85 & 41.10 \\

Cluster & \ding{55} 
& \cellcolor{lightgray}32.44 & \cellcolor{lightgray}37.48 
& 37.41 & 42.81 \\

Peaks & \ding{51} 
& \cellcolor{lightgray}36.54 & \cellcolor{lightgray}41.17 
& 42.23 & \textbf{48.67} \\

Cluster & \ding{51} 
& \cellcolor{lightgray}36.98 & \cellcolor{lightgray}41.10 
& \textbf{42.89} & 47.58 \\
\bottomrule
\end{tabular}%
}
\end{table}
%%%%%%%%%%%%%%%%%%%%%%%%%%%%%%
%                            %
%         Use Cases          %
%                            %
%%%%%%%%%%%%%%%%%%%%%%%%%%%%%%
\section{Use Cases}\label{sec:usecases}
\begin{figure*}[!t]
  \centering
  \includegraphics[width=.95\linewidth]{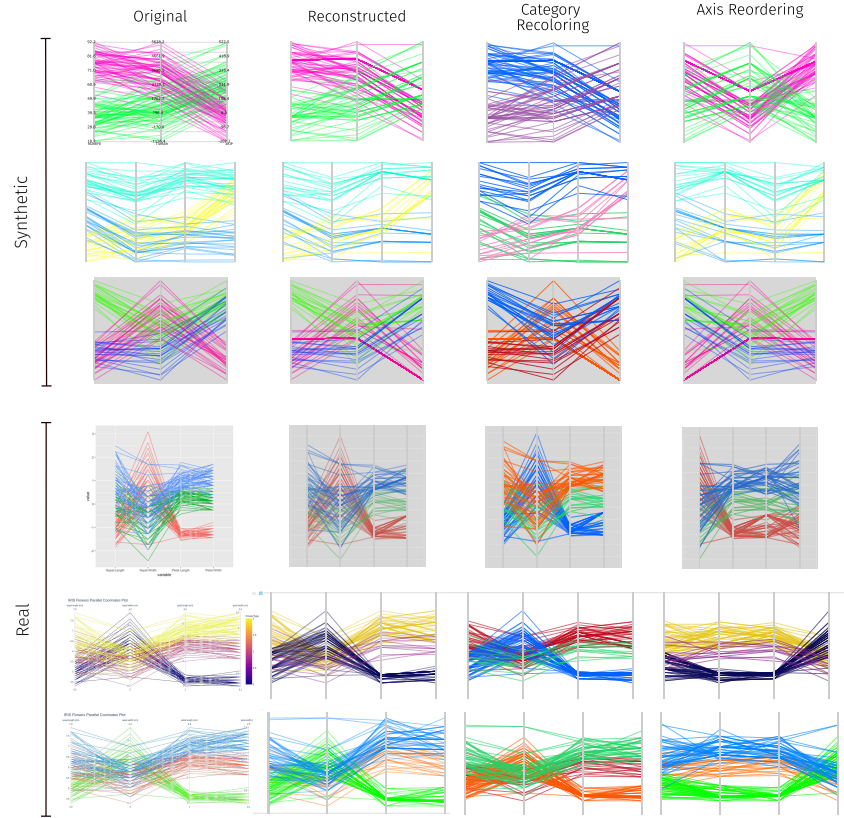}
  \caption{Applications enabled by PaCoNet using extracted datapoints. From left to right: original plots, reconstructions from predicted data, category recoloring, and axis reordering. Top three rows show synthetic data from our test dataset, bottom three rows show real parallel coordinate plots.}
   \label{fig:reconstruction}
\end{figure*}  
Beyond quantitative evaluation, PaCoNet enables post hoc visualization operations that are infeasible when working solely with raster images by exposing explicit datapoints. 
\textit{Chart reconstruction:} using extracted line coordinates and category assignments, parallel coordinates plots are re-rendered from predicted datapoints, producing editable visualizations that preserve consistent polyline identity across all axes.
\textit{Category recoloring:} recovered category memberships allow reconstructed plots to be recolored independently of the original encoding, enabling flexible palette selection and improved visual accessibility.
\textit{Axis reordering:} explicit datapoints further enable axis reordering by re-rendering polylines under alternative axis arrangements, facilitating the exploration of different variable relationships (~\autoref{fig:reconstruction}).
\\
Together, these examples demonstrate how PaCoNet transforms static parallel coordinates images into editable, data-driven representations. Further details are provided in the supplementary material.

%%%%%%%%%%%%%%%%%%%%%%%%%%%%%%
%                            %
%        Limitations         %
%                            %
%%%%%%%%%%%%%%%%%%%%%%%%%%%%%%
\section{Limitations}\label{sec:limitations}
PaCoNet is designed for standard parallel coordinates plots composed of straight polylines. More exotic variants, such as curved, surface-based, or heavily augmented parallel coordinates~\cite{graham2003using,mcdonnell2008illustrative,yuan2009scattering}, are outside the scope of this work and are not explicitly supported.
While we demonstrate qualitative generalization on real-world plots, the absence of quantitative evaluation on annotated real images remains a limitation, and constructing small-scale or weakly supervised benchmarks for real-world parallel coordinates is an important direction for future work.
The current axis detection pipeline assumes approximately linear, vertically aligned axes with regular spacing; plots with non-linear, inverted, or highly irregular layouts are not explicitly handled and may degrade performance.
PaCoNet does not recover semantic numeric values such as axis scales, tick labels, or textual annotations; instead, it serves as a structural preprocessing stage complementary to existing chart-mining pipelines, with OCR-based scale parsing left for future work.

%%%%%%%%%%%%%%%%%%%%%%%%%%%%%%
%                            %
%        Conclusions         %
%                            %
%%%%%%%%%%%%%%%%%%%%%%%%%%%%%%
\section{Conclusion}\label{sec:conclusions}
In this paper, we introduced PaCoNet, the first deep learning approach specifically designed for automated data extraction from parallel coordinates. We identify line extraction task grounded in literature, providing clear metrics for quantitative evaluation. Our newly created large-scale dataset, composed of synthetic and real-world parallel coordinate plots, enabled comprehensive training and evaluation, resulting in PaCoNet significantly outperforming baseline methods. This work not only advances automated data extraction from complex visualizations but also bridges the gap between computer vision and visualization research, paving the way for future investigations into advanced visualization analytics.

%%%%%%%%%%%%%%%%%%%%%  Paper ends %%%%%%%%%%%%%%%%%%%%%%%%%%%%%%%%%%%%%%%%%%%%

% \subsubsection{Acknowledgements} Please place your acknowledgments at
% the end of the paper, preceded by an unnumbered run-in heading (i.e.
% 3rd-level heading).


\bibliographystyle{splncs04}
\bibliography{mybibliography}





% LNCS Homepage, \url{http://www.springer.com/lncs}. Last accessed 4
% \end{thebibliography}
\end{document}

\title{Supplementary Material}

\author{Poonam Poonam\inst{1}\and
Hannah Kniesel\inst{1} \and
Pere-Pau Vázquez\inst{2} \and
Timo Ropinski\inst{1}}
\authorrunning{P. Poonam et al.}
\institute{Ulm University, Germany \\
\email{\{poonam.poonam, hannah.kniesel, timo.ropinski\}@uni-ulm.de}\\
Universitat Politècnica de Catalunya, Barcelona, Spain\\
\email{pere.pau.vazquez@upc.edu}
}
\maketitle             

This is supplementary material for our work "PaCoNet: Deep Data Extraction for Parallel Coordinates". 

\section{Real-World Dataset Collection and Curation}

To assemble a representative set of real-world parallel coordinates plots, we employed an automated web-scraping pipeline implemented in Python using the BeautifulSoup~\cite{richardson2007beautiful} library. Images were retrieved from publicly accessible web sources via Google and Bing image search using query terms such as \textit{parallel coordinates} and \textit{no angular parallel coordinates}. 

Since automated scraping yields heterogeneous results, all collected images were manually curated. We removed non-parallel-coordinate visualizations, duplicates, and plots with severe rendering artifacts or insufficient visual quality. This curation process ensured that the retained images reflect realistic and interpretable examples of parallel coordinates plots as encountered in practice. The final real-world dataset consists of approximately 200 images.

\section{Real-World Statistics Extraction Protocol}

We analyze the real-world dataset to estimate key visual and structural properties of parallel coordinates plots used to guide synthetic data generation.

\paragraph{Axes.}
The number of vertical axes is determined by manually counting visible axes in each plot, which are clearly identifiable due to their prominent vertical alignment.

\paragraph{Data Density.}
The number of data rows (polylines) is estimated using density-based binning. Each plot is assigned to a predefined density bin (from sparse to highly dense), avoiding unreliable exact counting in cases of heavy overdraw while still capturing meaningful variation in data scale.

\paragraph{Categories.}
The number of categories is estimated by identifying visually distinct color groups used for categorical encoding. Plots without explicit categorical encoding are assigned a single category.

\paragraph{Grid, Ticks, and Labels.}
The presence of grid lines, axis ticks, and textual labels is recorded as binary attributes based on visual inspection.

\paragraph{Background Appearance.}
To characterize background color properties, background pixels are sampled from visually homogeneous regions of each image, and the mean RGB intensity is computed.

All extracted attributes are aggregated into histograms or bar charts, as shown in Figure~2 of the main paper, and are used to guide the parameter ranges and design choices of the synthetic data generator.

\section{Synthetic Data Generator Details}

While the real-world dataset provides valuable insights into the structure and appearance of parallel coordinates plots, its limited size makes it unsuitable for training a deep neural network. More importantly, these images lack pixel-level and structural ground-truth annotations, which prevents their direct use for supervised learning. To enable supervised learning at scale, we therefore generate a large synthetic dataset of parallel coordinates plots with exact ground-truth annotations.
The design of the synthetic data generator is informed by the statistical analysis of the real-world dataset. Rather than generating plots with arbitrary parameters, we sample key properties from distributions that match those observed in the obtained real-world visualizations. This strategy ensures that the synthetic data captures realistic variations while remaining fully controllable and precisely annotated.
\subsection{Structural Parameters}

The number of vertical axes, data rows, and categories for each synthetic parallel coordinates plot is sampled from empirical distributions estimated from the real-world dataset. These distributions capture both common cases with a small to moderate number of axes and rows, as well as less frequent but more complex plots with higher dimensionality and data density. Sampling from these distributions ensures coverage of a wide range of structural configurations observed in practice.

\subsection{Visual Appearance}

Visual attributes are sampled to reflect real-world rendering styles. These include the presence or absence of grid lines, axis ticks, and labels, as well as background brightness and line color distributions. Background color is sampled using the empirical RGB intensity distribution observed in real-world plots, while line colors are sampled to reflect common categorical color encodings. Unless constrained by empirical observations, visual attributes are sampled independently to maximize variability.

\subsection{Data Generation and Rendering}

Synthetic parallel coordinates plots are generated using the Vega-Lite visualization grammar~\cite{satyanarayan2016vega}. Vega-Lite specifications define axis placement, polyline rendering, color encoding, and optional visual elements such as grids and labels. All rendering parameters and configuration settings are fixed across experiments to ensure reproducibility. 
\paragraph{Image Resolution.}
All synthetic parallel coordinates plots are rendered at a fixed resolution of $600 \times 300$ pixels. This resolution reflects common aspect ratios and chart sizes observed in real-world parallel coordinates plots and is used consistently across training, validation, and testing.

\subsection{Ground-Truth Annotations}

Since all plots are generated programmatically using Vega-Lite visualization Grammar~\cite{satyanarayan2016vega}, complete ground-truth annotations are retained by construction. These include exact axis locations, underlying data values, and polyline correspondences across axes. This enables supervision at a level of precision that is infeasible to obtain from real-world raster images.

\section{Additional Experimental Details}

\subsection{Category Separation Metrics}

This section provides detailed definitions of the metrics used to evaluate category separation design choices.

\paragraph{Silhouette Coefficient (SC).}
The Silhouette Coefficient~\cite{lovmar2005silhouette} measures how well individual samples are assigned to clusters by comparing intra-cluster compactness with inter-cluster separation. Given a pixel $i$, its silhouette score is defined as
\[
s(i) = \frac{b(i) - a(i)}{\max(a(i), b(i))},
\]
where $a(i)$ is the average distance to other samples within the same cluster and $b(i)$ is the minimum average distance to samples in other clusters. Higher values indicate better-separated and more coherent category groupings.

\paragraph{Category Count MAE (CC-MAE).}
To assess the accuracy of category count estimation, we report the Mean Absolute Error (MAE)~\cite{willmott2005advantages} between the predicted and ground-truth number of categories for each plot.

\paragraph{Intersection-over-Union (IoU).}
When pixel-level category assignments are available, we evaluate spatial consistency using Intersection-over-Union (IoU)~\cite{Rezatofighi_2019_CVPR} between predicted and ground-truth category masks. IoU is particularly informative in regions with heavy line overlap or dense crossings.

\subsection{Additional Category Separation Analysis}

\paragraph{Color Space Selection.}
We evaluated RGB, LAB, and HSV color spaces for category separation. While RGB and LAB occasionally separate categories under high contrast, the hue component of HSV consistently produces well-separated, unimodal color clusters corresponding to categorical encodings. This observation motivates our use of hue-only representations in the main pipeline.

\paragraph{Image Resolution Effects.}
We observe that clustering-based separation benefits from moderate downscaling of the input image, which reduces pixel-level noise and stabilizes density estimation. In contrast, peak-based separation relies on high-resolution hue histograms and therefore performs best at full resolution.

\paragraph{DBSCAN vs. HDBSCAN.}
We compared DBSCAN~\cite{ester1996density} and HDBSCAN~\cite{mcinnes2017hdbscan} for clustering-based category separation. DBSCAN consistently outperforms HDBSCAN in our setting due to the relatively uniform, one-dimensional hue feature space, where categories form well-separated clusters with similar density. HDBSCAN’s hierarchical processing, while beneficial in variable-density settings, introduces unnecessary complexity in this regime.

\subsection{Vision-Language Model (VLM) Prompting Details}

We include large vision-language models (VLMs) as reference baselines to reflect current user practice for chart understanding. Since these models do not natively output structured geometric representations, we prompt them to regress explicit line coordinates in pixle space.

\paragraph{Prompting Strategy.}
Each VLM is provided with the input parallel coordinates image and instructed to identify all visible data polylines. The model is prompted to output the start and end coordinates of each detected line segment in image pixel coordinates. To ensure parsable outputs, responses are constrained to a structured format listing line endpoints.

\paragraph{Output Parsing and Evaluation.}
The predicted line coordinates are parsed automatically and evaluated using LC-MAE, measuring discrepancies between predicted and ground-truth line coordinates and counts. No post-processing or manual correction is applied to VLM outputs.

\paragraph{Limitations of VLM Baselines.}
We emphasize that VLMs are not designed for precise, pixel-level geometric extraction, particularly in densely overdrawn settings such as parallel coordinates plots. Their inclusion serves to contextualize PaCoNet’s performance relative to general-purpose multimodal models rather than as a task-specific baseline.

\paragraph{Vision-Language Model Prompt.}
The following prompt was used verbatim to obtain line-level predictions from vision-language models. The prompt enforces a strict output format to prevent free-form reasoning and to enable automatic parsing for quantitative evaluation.

\begin{tcolorbox}[
  colback=gray!8,
  colframe=gray!60,
  boxrule=0.6pt,
  arc=3pt
]
\footnotesize
\begin{verbatim}
Extract all visible straight line segments (plotted data lines) 
in the image.

Return ONLY valid JSON in exactly this format:
{
  "lines": [
    [x0, y0, x1, y1],
    ...
  ]
}

Rules:
- No markdown, no extra text, no explanations.
- Coordinates are IMAGE PIXELS with origin at top-left.
- Ignore axes, tick marks, grid lines, legend boxes, and text.
- Use numbers (integers or floats).
- Lines belonging to each semantic color category in the image.
\end{verbatim}
\end{tcolorbox}

\section{Detailed Use Case Analysis}

Beyond quantitative evaluation, the datapoints extracted by PaCoNet enable a range of downstream visualization operations that are not feasible when working with raster images alone. By recovering explicit polyline geometry and category assignments, PaCoNet converts static parallel coordinates plots into editable, data-driven representations.

\paragraph{Chart Reconstruction.}
PaCoNet enables the reconstruction of parallel coordinates plots directly from predicted datapoints. Using the extracted line coordinates and category assignments, polylines are re-rendered between estimated axis positions without access to the original underlying dataset. Despite operating solely on image-derived predictions, the reconstructed plots closely match the visual structure of the originals. This demonstrates that PaCoNet preserves consistent polyline identity across all axes rather than producing independent local segments, which is essential for faithful reconstruction and subsequent data manipulation.

\paragraph{Category Recoloring.}
Because category membership is explicitly recovered during the extraction process, reconstructed plots can be recolored independently of the original visual encoding. This allows users to apply alternative color palettes, improve contrast, or adapt visualizations for accessibility requirements such as color-vision deficiencies. Importantly, recoloring operates directly on the reconstructed datapoints, enabling flexible visual redesign without reprocessing the original raster image.

\paragraph{Axis Reordering.}
Access to explicit datapoints further enables structural modifications of the visualization. In particular, vertical axes can be reordered to explore alternative variable arrangements and reveal different correlation patterns. Axis reordering is achieved by re-rendering the reconstructed polylines under a new axis order, ensuring that polyline continuity and category assignments are preserved. This operation highlights how PaCoNet supports exploratory analysis workflows that are otherwise inaccessible from static images.

Together, these use cases demonstrate that PaCoNet goes beyond passive image interpretation and enables interactive, post hoc analysis of parallel coordinates plots by exposing their underlying geometric structure.

\bibliographystyle{splncs04}
\bibliography{mybibliography}